\pdfoutput=1
\documentclass[11pt]{article}

\usepackage{acl}

\usepackage{times} 
\usepackage{latexsym}

\usepackage[T1]{fontenc}
\usepackage[utf8]{inputenc}

\usepackage{microtype}

\usepackage{booktabs}
\usepackage{graphicx}
\usepackage{stfloats}
\usepackage{multirow} 
\usepackage{color}
\usepackage{color}

\title{NLNDE at SemEval-2023 Task 12: Adaptive Pretraining and Source Language Selection for Low-Resource Multilingual Sentiment Analysis}

\author{Mingyang Wang$^{1,2}$, %\And
	Heike Adel$^1$, %\And
        Lukas Lange$^{1}$, %\AND
	{\bf Jannik Str\"{o}tgen}$^{1,3}${\bf,} {\bf Hinrich Sch\"{u}tze}$^2$ \\
	\hspace{0cm}$^1$ Bosch Center for Artificial Intelligence, Renningen, Germany\\
	\hspace{0cm}$^2$ Center for Information and Language Processing (CIS), LMU Munich, Germany \\
     \hspace{0cm}$^3$ Karlsruhe University of Applied Sciences, Karlsruhe, Germany\\
	{\tt \hspace{0cm}\{Mingyang.Wang2,Heike.Adel,Lukas.Lange\}@de.bosch.com} \\
}

\begin{document}
\maketitle
\begin{abstract}
This paper describes our system developed for the SemEval-2023 Task 12 ``Sentiment Analysis for Low-resource African Languages using Twitter Dataset''. Sentiment analysis is one of the most widely studied applications in natural language processing. However, most prior work still focuses on a small number of high-resource languages. Building reliable sentiment analysis systems for low-resource languages remains challenging, due to the limited training data in this task. In this work, we propose to leverage language-adaptive and task-adaptive pretraining on African texts and study transfer learning with source language selection on top of an African language-centric pretrained language model.
Our key findings are: (1) Adapting the pretrained model to the target language and task using a small yet relevant corpus improves performance remarkably by more than 10 F1 score points. (2) Selecting source languages with positive transfer gains during training can avoid harmful interference from dissimilar languages, leading to better results in multilingual and cross-lingual settings. In the shared task, our system wins 8 out of 15 tracks and, in particular, performs best in the multilingual evaluation.
\end{abstract}

\section{Introduction}
In recent years, natural language processing research has attracted considerable interest. However, most studies remain confined to a small number of languages with large amounts of training data available. Low-resource languages, 
for example, e.g., African languages, are still underrepresented although they are spoken by over a billion people. In this context, the AfriSenti shared task provides a Twitter dataset for sentiment analysis on 14 African languages, promoting the future development of this field. The shared task consists of three sub-tasks: monolingual (Subtask A), multilingual (Subtask B), and zero-shot cross-lingual sentiment analysis (Subtask C). A detailed description can be found in the shared task description papers \cite{muhammad2023afrisenti, muhammadSemEval2023}.

In this paper, we describe our submission as Neither Language Nor Domain Experts (NLNDE)\footnote{We neither know any African languages nor have prior knowledge of the Twitter domain dataset.} to the AfriSenti shared task. Given the key challenge of limited training data,  
we first adopt the language-adaptive and task-adaptive pretraining approaches \cite{dont-stop}, i.e., LAPT and TAPT,  to adapt a pretrained language model to the language and task of interest. Further pretraining the model with such smaller but more task-relevant corpora leads to performance gains in all subtasks. 

Second, Cross-lingual transfer has been shown to be an effective method for enhancing the performance of low-resource languages by leveraging one or more similar languages as source languages \cite{lin-etal-2019-choosing, ruder-plank, nasir-mchechesi-2022-geographical}. However, the 14 African languages covered in this shared task (see Table \ref{tab:language-info}) come from different language families and, therefore, hold different linguistic characteristics. As dissimilar languages could hurt the transfer performance \cite{lin-etal-2019-choosing, masakhaner2}, it is important to choose promising languages as the transfer source. Therefore, our system uses transfer learning with an explicit selection of source languages.
We apply this approach to the multilingual and zero-shot cross-lingual sentiment analysis tasks (Subtask B and C) and demonstrate that it benefits the performance of each target language. In addition, we investigate different source language selection strategies and show their impact on the final transfer performance.

Our final submission results are the ensemble of the best models with different random seeds. Our system is ranked first in 6 out of 12 languages in subtask A (monolingual), achieves the first place in subtask B (multilingual), and wins
for one of two languages
%in the Tigrinya language 
in subtask C for the zero-shot cross-lingual transfer.

\section{System Overview}

Our system is based on the AfroXLM-R large model \cite{afroxlm-r}, which applys multi\-lingual adaptive fine-tuning on XLM-R \cite{xlm-r} with a special focus on African languages. In all three subtasks, we first apply language- and/or task-adaptive pretraining with language- and/or task-specific data to tailor the vanilla AfroXLM-R to our setting. In subtasks B and C, after the adaptive pretraining, we perform source language selection to improve the multilingual and cross-lingual transfer performance.

\subsection{Language- and Task-Adaptive Pretraining}
\label{sec:adaptative pretraining}
Most of the current NLP research is based on large language models that have been pretrained on massive amounts of heterogeneous corpora. \newcite{dont-stop} demonstrate that it is helpful to further tailor a pretrained model to the domain of the target task. They show that continued pretraining with domain-specific and task-specific data consistently improves performance on domain-specific tasks across different domains and tasks. Specifically, they introduce \emph{domain-adaptive pretraining} (DAPT), i.e., the continued pretraining of the base model on a large corpus of unlabeled domain-specific text. Analogously, \emph{task-adaptive pretraining} (TAPT) refers to adapting the pretrained model to the task’s unlabeled training data.

A natural extension is the application of this method to multilingual scenarios. Considering different languages as different domains, language-adaptive pretraining can be viewed as a special case of domain-adaptive pretraining. Therefore, we also explore two types of continued pretraining: First, we pretrain the base model with a language-specific corpus, which we will refer to as \emph{language-adaptive pretraining} (LAPT).\footnote{In some 
%literature 
works
\cite{afro-lm, afroxlm-r}, LAPT is also called language-adaptive fine-tuning (LAFT).} Second, we  adapt the language model on the unlabeled task dataset, i.e., perform \emph{task-adaptive pretraining} (TAPT). 
%, it also refers to the continued training of PLMs on the language corpus with the pretraining objective. We will use the term LAPT in the following sections to be consistent with \cite{dont-stop}.

For the language-specific pretraining, we collect open-source corpora from the multi-domain Leipzig Corpus Collection \cite{leipzig-corpus}, covering Wikipedia, Community, Web, and News corpora. Note that the final set of monolingual corpora depends on their availability. There is not a single corpus covering all these languages. Table \ref{tab:language-info} provides a summary of the monolingual corpora we used for our language-adaptive pretraining.

\begin{table*}[htbp]
  \centering
  \footnotesize
  \scalebox{0.8}{
    \begin{tabular}{lllrrll}
    \toprule
    \textbf{Language} & \textbf{Family} & \textbf{LAPT Source} & \multicolumn{1}{l}{\textbf{Size (MB)}} & \multicolumn{1}{c}{\textbf{No. of Sent}} & \textbf{Fwd. source} & \textbf{Bwd. source} \\
    \midrule
    \multicolumn{7}{l}{\textit{Subtask A and B}} \\
    Amharic (am) & Afro-Asiatic / Semitic & Wikipedia & 17    & 12,861 & dz, ha, kr, ma & kr, ma \\
    Algerian Arabic (dz) & Afro-Asiatic / Semitic & News  & 15.3  & 59,998 & ha, kr, ma & kr, ts, twi, ma, am \\
    Hausa (ha) & Afro-Asiatic / Chadic & Wikipedia & 6.5   & 59,664 & kr, twi, dz, pcm & kr, ts, twi, dz, am \\
    Igbo (ig) & Niger-Congo / Volta-Niger & Wikipedia & 2.2   & 17,786 & sw, pcm, dz, yo & kr. ts, am, ma, dz \\
    Kinyarwanda (kr) & Niger-Congo / Bantu & Community & 7.9   & 60,480 & -      & ma, am, ts, dz, twi \\
    Moroccan Arabic (ma) & Afro-Asiatic / Semitic & News  & 7.4   & 29,997 & yo, dz, pcm, ha & kr, ts, am, dz, twi \\
    Nigerian Pidgin (pcm) & English-Creole & Community & 0.4   & 7,126 & dz    & kr, ts, twi, dz, ma \\
    Mozambican Portuguese (pt) & Indo-European & Web   & 27    & 197,340 & dz, yo, ma, am & kr, ma, am \\
    Swahili (sw) & Niger-Congo / Bantu & Wikipedia & 3.2   & 28,945 & yo, dz, pcm, ha & ma, ts, yo, pcm \\
    Xitsonga (ts) & Niger-Congo / Bantu & Web, Community & 3.6   & 30,513 & ha, ma, pcm, pt & kr, twi, pcm, dz \\
    Twi (tw) & Niger-Congo / Kwa & Wikipedia & 3.4   & 2,478 & ig, kr & kr, ts, pcm, dz \\
    Yoruba (yo) & Niger-Congo / Volta-Niger & Wikipedia, Community & 4.5   & 41,291 & -    & kr, ts, twi, dz \\
    \midrule
    \multicolumn{7}{l}{\textit{Subtask C}} \\
    Oromo (or) & Afro-Asiatic / Cushitic &       &       &       & kr, ha, yo, ts & yo, pt, ts \\
    Tigrinya(tg) & Afro-Asiatic / Semitic &       &       &       & ha, kr, am, ma, pt & pt, yo, ha \\
    \bottomrule
    \end{tabular}%
    }%
  \caption{Language information, pretraining corpora statistics and forward/backward source language selection results. Source corpora for language-adaptive pretraining(LAPT) sources are collected from Leipzig Corpus Collection \cite{leipzig-corpus}. The source language selection process is described in Section \ref{sec:source selection}. The selection is based on the weighted F1 as transfer score averaged over 5 random seeds.}
  \label{tab:language-info}%
\end{table*}%

\subsection{Transfer Learning and Source Selection}
\label{sec:source selection}

Table \ref{tab:language-info} provides an overview of the languages covered in the shared task. They come from four 
%different 
language families (Afro-Asiatic, Niger-Congo, English-Creole and Indo-European) and, therefore, have different linguistic characteristics. Even inside the same language family, languages can still exhibit distinct linguistic features. For example, although many languages are from the Niger-Congo family (6 out of 14 languages), affixes are very common in Bantu languages (a subgroup of Niger-Congo), but are not typically used in non-Bantu subgroups like Volta-Niger and Kwa.

Previous work has demonstrated that it can still be beneficial to leverage one or more similar languages for cross-lingual transfer learning to the target language \cite{lin-etal-2019-choosing}. Nevertheless, languages that are dissimilar to the target language could also hinder performance \cite{masakhaner2, to-share-or-not}. Therefore, it is crucial to properly select source languages to improve the transfer results on the target language. 

In this work, we use transfer learning with selected sources for Subtask B and C, as they both involve transferring from multiple languages to a target language. For source language selection, we perform forward and backward source language selection, similar to the corresponding feature selection approaches \cite{fwd-bwd-1, fwd-bwd-2}.\footnote{also known as variable selection.} Essentially, feature selection is defined as the problem of selecting a minimal-size subset of features that leads to an optimal, multivariate predictive model for a target of interest \cite{fwd-bwd-1}. In our task, we consider each candidate source language as a feature. For each target language, we aim to filter out irrelevant or harmful source languages and only keep the beneficial languages as the transfer source.

\emph{Forward feature selection} usually starts with an empty set of features and adds variables to it, while \emph{backward feature selection} starts with a  complete set of variables and then excludes variables from it. In particular, for forward language selection, given a target language $L_{t}$, we start with a set $S_{fwd} = \{L_{t}\}$ containing only the target language. We then add each of the other languages $L_{s_{i}}, i=1\ldots N-1$ at a time and obtain $N-1$ bilingual sets $\{( L_{t}, L_{s_{i}})\}_{i=1\ldots N-1}$, each with the target language $L_{t}$ and one source language $L_{s_{i}}$. $N$ refers to the total number of 
%the 
given languages. We experiment with each bilingual language set to build the training dataset for transfer learning. 
Additionally, we run $N$ monolingual experiments (one per target language) and use the monolingual performance as the baseline to determine if a candidate source language leads to positive or negative transfer gains. To be more specific: If a bilingual language set $(L_{t}, L_{s_{i}})$ yields a score of more than 5\% above the monolingual performance with $L_{t}$, we 
%then 
consider $L_{s_{i}}$ as a positive source with respect to the target language $L_{t}$. 
%In the following sections, we will use the term \textit{positive source languages} to refer to the source languages that yield positive transfer with respect to a target language.

For backward selection, we start with the complete language set with all $N$ languages. For each target language, we exclude each of the other $N-1$ languages and get $N-1$ language sets, denoted as $\{( L_{t}, L_{s_{1}}\ldots L_{s_{i-1}}, L_{s_{i+1}}\ldots L_{s_{N-1}})\}_{i=1\ldots N-1}$. To get a baseline performance for comparison, we randomly select 500 samples from each language to build a small multilingual set. We choose a constant number of samples per language to avoid side effects of the data size on the performance. Given the set of all languages, we remove each language at a time and compare it with the baseline results from the complete language set to investigate the transfer gain of each candidate language on the final performance. If the performance from the language set $(L_{t}, L_{s_{1}}\ldots L_{s_{i-1}}, L_{s_{i+1}}\ldots L_{s_{N-1}})$ is more than 5\% below the baseline, it shows that the absence of $L_{s_{i}}$ has a large negative impact on performance. %Therefore, $L_{s_{i}}$ is considered as a positive source language to the target $L_{t}$.

For each of the $N$ languages, we need to run $N-1$ experiments with the bilingual language set and $1$ baseline experiment with the monolingual dataset. Therefore, for forward source language selection, we need to run $N \times N$ transfer experiments and then select the source languages with positive transfer gains corresponding to each target language via the performance comparison. Similarly, for backward source language selection,  $N \times N$ experiments are required for the source language selection.

We apply both selection strategies for subtasks B and C. In subtask C, the language sets do not contain the target language $L_{t}$ (as it is a zero-shot task). In particular, for forward selection, this means that we start with an empty set. For the same reason, We run experiments with the complete datasets of all languages as the baseline for both forward and backward selection, as there is no monolingual dataset for the target language in subtask C.  Results for the source language selection of Subtask B and C are given in Table \ref{tab:language-info}.

\section{Experimental Setup}
We now provide details on our preprocessing steps, the language models and their training. 

\subsection{Data Preprocessing}
\label{sec:data_preprocess}
We preprocess the raw input tweets by removing extra whitespaces, incorrect repeated characters and punctuation. Similar to \newcite{bertweet},
%\cite{bertweet}, 
we replace all URLs with “HTTPURL” and username mentions with “USER” as they have little to no impact on sentiment analysis. When analyzing the data, we noticed a small portion of samples overlapped in the train and dev sets for some languages. Therefore, to measure the actual generalizability of our models, we remove all the overlapping samples from the dev set. We will use \textit{dev set*} to denote the processed dev set in the following.

\begin{table*}[htbp]
  \centering
  \footnotesize
  \scalebox{0.9}{
    \begin{tabular}{lccccccccccccc}
    \toprule
    \multicolumn{1}{c}{\textbf{Model}} & \textbf{am} & \textbf{dz} & \textbf{ha} & \textbf{ig} & \textbf{kr} & \textbf{ma} & \textbf{pcm} & \textbf{pt} & \textbf{sw} & \textbf{ts} & \textbf{twi} & \textbf{yo} & \textbf{average} \\
    \midrule
    AfroXLM-R & 31.60 & 60.95 & 80.78 & 68.82 & 70.34 & 43.15 & 48.27 & 62.60 & 58.96 & 47.39 & 36.59 & 55.67 & 62.91 \\
    \midrule
    \multicolumn{14}{l}{\textit{AfroXLM-R with adaptive pretraining}} \\
    LAPT  & 47.26 & 66.48 & 79.53 & 80.40 & 70.15 & 48.16 & 67.22 & 69.29 & 62.66 & 57.92 & 56.81 & 76.39 & 69.56 \\
    TAPT  & 61.92 & \textbf{68.28} & \textbf{81.61} & \textbf{81.49} & 70.40 & \textbf{63.58} & \textbf{70.08} & \textbf{71.07} & 63.38 & 38.96 & \textbf{67.10} & 78.03 & \textbf{73.49} \\
    LAPT + TAPT & \textbf{63.87} & 67.63 & 80.74 & 81.45 & \textbf{70.67} & 61.7  & 69.91 & 70.66 & \textbf{64.04} & \textbf{59.49} & 66.49 & \textbf{78.57} & 73.23 \\
    \midrule
    \multicolumn{14}{l}{\textit{Submitted systems}} \\
    king001 & 69.77 & \textbf{73.00} & 81.11 & 81.39 & 60.26 & 57.94 & \textbf{75.75} & 73.53 & 64.89 & \textbf{68.28} & 56.26 & \textbf{80.16} & 73.82 \\
    PALI  & 65.56 & 72.62 & 81.10 & 81.30 & 69.61 & 55.92 & 75.16 & \textbf{73.83} & 64.37 & 67.58 & 56.26 & 80.06 & 73.64 \\
    stce  & 65.56 & 71.72 & 80.99 & 81.37 & 69.61 & 55.42 & 75.30 & 73.57 & 64.37 & 67.58 & 56.26 & 80.08 & 73.56 \\
    UM6P  & \textbf{72.18} & 72.02 & 82.04 & 81.51 & 70.71 & 60.15 & 69.14 & 67.35 & 60.26 & 66.98 & 56.13 & 76.01 & 73.31 \\
    \textbf{NLNDE (ours)} & 64.04 & 69.98 & \textbf{82.62} & \textbf{82.96} & \textbf{72.63} & \textbf{64.82} & 71.93 & 72.90 & \textbf{65.67} & 60.70 & \textbf{67.51} & 79.95 & \textbf{74.78} \\
    \bottomrule
    \end{tabular}%
    }
  \caption{Performance on subtask A: Monolingual sentiment analysis. We apply LAPT, TAPT and a combination of both on top of the AfroXLM-R model. All adaptive pertraining methods significantly improve performance. TAPT achieves the best overall F1 score, and also the best on 7 out of 12 languages. We calculate the average F1 scores of each model (weighted based on the number of samples of each language). For the submission as NLNDE, we ensemble different random seeds of all three models with adaptive pretraining and achieve even better performance.}
  \label{tab:task_a_results}%
\end{table*}%

\subsection{Pretrained Language Models} 
Large multilingual pretrained language models (PLMs) like mBERT \cite{bert} and XLM-R \cite{xlm-r} have shown impressive capability on many languages for a variety of downstream NLP tasks. They are also often used as initialization checkpoints for adapting to other languages, such as AfroXLM-R, which is initialized from XLM-R and specialized to African languages. In initial experiments, we compare the performance of several multilingual PLMs, including BERT and XLM-R which are trained on hundreds of languages, and AfroLM \cite{afro-lm}, Afro-XLM-R  \cite{afroxlm-r} as African language-specific models. AfroXLM-R performs best across all three subtasks. Therefore, we select AfroXLM-R large as our base model and apply adaptive pretraining and source language selection on top of it. In addition, in subtask C, we experiment with translating the tweets into English and apply BERTweet \cite{bertweet}, a pretrained language model for English tweets.

\subsection{Training Details} 
%As mentioned above, we use AfroXLM-R as our base model and further apply adaptive pretraining and source language selection on top of it. 
For task- and language-adaptive pretraining, we use the AdamW optimizer \cite{adamw} with a learning rate of 5e-5 and a batch size of 8. For fine-tuning, we use Adam with a learning rate of 2e-5 and a batch size of 32. In both phases, we use a maximum sequence length of 128. The training was done on Nvidia A100 and V100 GPUs.\footnote{All experiments ran on a carbon-neutral GPU cluster.} The results are evaluated using the weighted F1 score on the test set averaged over 5 random seeds. The final submission comes from the majority vote ensemble of different random seeds of the best models.

\section{Results}
In this section, we report our results on the three subtasks and discuss our findings and observed limitations of the current work. Our evaluation is based on the weighted F1 score on the test set averaged over 5 random seeds. We use the majority vote method to ensemble our models from different random seeds for submission, we provide the final submission results, as well as the results from several top-ranked systems in the last lines in Table \ref{tab:task_a_results} $\sim$ \ref{tab:task_c_results}. We refer to Appendix \ref{sec:result_dev_set} for the results on the \textit{development set}.

% Our submitted system is denoted by our team name NLNDE (neither language nor domain experts), following previous successful participations in NLP shared tasks for non-standard languages and domains \cite{lange-etal-2019-pharmaconer-system,lange-etal-2019-meddocan-system,lange-etal-2020-cantemist-system,lange-etal-2021-meddoprof-system}. % TODO: back in crv

\subsection{Subtask A: Monolingual Sentiment Analysis}
\label{sec:results_subtask_a}

In subtask A, we mainly study the impact of adaptive pretraining on monolingual sentiment analysis. We use the off-the-shelf AfroXLM-R large model as our baseline and fine-tune it on the training dataset of each language, yielding one fine-tuned model per language. Then, we apply LAPT, TAPT and their combination on top of AfroXLM-R. For combined LAPT and TAPT, we begin with AfroXLM-R and apply LAPT then TAPT for the model adaptation. After pretraining, we fine-tune the adapted model on each monolingual dataset for sentiment analysis.

As shown in Table \ref{tab:task_a_results}, the performance is remarkably improved with adaptive pretraining for most languages, especially with task-adaptive pretraining (TAPT), which leads to a performance gain of 10.58 F1 score on average. LAPT also increases the performance in general, but does not contribute that much in comparison and even degrades the performance for the languages Hausa (ha) and Kinyarwanda (kr). We speculate that, on the one hand, we use relatively small language-specific corpora for LAPT as the covered African languages are indeed low-resource. In contrast, \newcite{dont-stop} used much larger adaptation corpora for domain-specific pretraining (DAPT). On the other hand, the mismatch of text domains might be another reason: As described in Section \ref{sec:adaptative pretraining}, we use corpora from domains, such as news and Wikipedia for LAPT, while the actual task dataset consists of multilingual tweets involving many Twitter-specific factors, such as code-mixing, misspellings, emojis, or hashtags.

Combining LAPT and TAPT also shows promising results. However, as analyzed before, we hypothesize that most of the benefits come from the more effective TAPT.

\subsection{Subtask B: Multilingual Sentiment Analysis}
\label{sec:results_subtask_b}

\begin{table}[!tbp]
\footnotesize
  \centering
    \begin{tabular}{lc}
    \toprule
    \multicolumn{1}{c}{\textbf{Model}} & \textbf{Overall F1} \\
    \midrule
    %\multicolumn{2}{l}{\textit{Multilingual training}} \\
    Multilingual & 48.41 \\
    + TAPT & 70.65 \\
    \midrule
    %\multicolumn{2}{l}{\textit{Monolingual training}} \\
    Monolingual & 62.91 \\
    + LAPT & 69.56 \\
    + TAPT & 73.49 \\
    + LAPT \& TAPT & 73.23 \\
    \midrule
    \multicolumn{2}{l}{\textit{Transfer with selected sources}} \\
    Language family grouping & 61.81\\
    Fwd source transfer  & 66.73 \\
    Fwd source transfer + TAPT & 73.50 \\
    Bwd source transfer  & 66.41 \\
    Bwd source transfer + TAPT & \textbf{74.08} \\
    \midrule
    \multicolumn{2}{l}{\textit{Submitted systems}} \\
    king001 & 74.96 \\
    DN    & 72.55 \\
    ymf924 & 72.34 \\
    mitchelldehaven & 72.33 \\
    \textbf{NLNDE (ours)} & \textbf{75.06} \\
    \bottomrule
    \end{tabular}%
  \caption{Performance on subtask B: Multilingual sentiment analysis. We conduct (1) multilingual (2) monolingual and (3) transfer experiments. The single model with TAPT and backward language selection achieves the best overall results with an F1 score of 74.08. For the submission as NLNDE, we ensemble different random seeds of models with source selection and TAPT (forward and backward) and get the final results.}
  \label{tab:task_b_results}%
\end{table}%

In the multilingual subtask, we categorize our experiments into three groups: (1) multilingual training of a single model, (2) monolingual training of language-specific models and (3) transfer learning with selected sources. They differ in the composition of the training datasets. In multilingual training, we use all training data from 12 languages. In monolingual training, we use the same language-specific models as in Subtask A (see Section \ref{sec:results_subtask_a}) and combine the predictions in the end.  In transfer learning with selected sources, we perform forward and backward source selection as described in Section \ref{sec:source selection}. With the selected source languages given in Table \ref{tab:task_b_results}, we build the respective training datasets and fine-tune the model for each language. As a baseline, we further group languages based on their language family. This results in four groups,
% coming from four language families, 
namely Afro-Asiatic, Niger-Congo, English-Creole and Indo-European, details are given in Table \ref{tab:language-info}.

Our multilingual sentiment analysis results are given in Table \ref{tab:task_b_results}. First, 
% same 
as in Subtask A, task-adaptive pretraining notably improves classification performance in all task settings. Combining LAPT and TAPT is not better than TAPT only.
Therefore, due to time constraints, we apply LAPT and LAPT+TAPT only in monolingual training, but not in multilingual training and the transfer learning with selected sources. 

Second, fine-tuning the model individually to each target language (monolingual training) outperforms the joint multilingual training, in both the vanilla training (62.91 vs.\ 48.41) and adaptive pretraining (73.49 vs.\ 70.65) cases. Furthermore, selecting source languages with positive gains for each target language can further enhance performance over monolingual training. Grouping languages based on their language families (61.81) shows better results than multilingual training (48.41), but it underperforms monolingual training (62.91) and falls behind the transfer learning with selected sources (66.73 and 66.41) by around 5\%. %This demonstrates that our approach for source language selection leads to better cross-lingual transfer than the intuitive grouping purely based on the language family.} 
Forward and backward source selection gives different results, but they both contribute to the final results.

Finally, another interesting finding is that, in the presence of TAPT, the advantage of specifying languages as training data, i.e., in the cases of monolingual training and transfer learning with selected sources, becomes less pronounced. Specifically, without TAPT, multilingual training achieves an F1 score of 48.41, while monolingual training achieves 62.91 and transfer learning with selected sources achieves 66.73 and 66.41. However, with TAPT, the multilingual training shows a large improvement, yielding an F1 score of 70.65. Although monolingual (73.49) and transfer with selected languages (73.50 and 74.08) still outperform the multilingual result, they become less advantageous. We suppose this is because, with task-adaptive pretraining, the model already adapts to the target language compared with the vanilla model pretrained on a larger language set. As a result, the effect of additionally specifying the source languages is limited.

\subsection{Subtask C: Zero-shot Cross-Lingual Sentiment Analysis}
\label{sec:results_subtask_c}

\begin{table}[!tbp]
\footnotesize
  \centering
    \begin{tabular}{lcc}
    \toprule
    \multicolumn{1}{c}{\textbf{Model}} & \textbf{or} & \textbf{tg} \\
    \midrule
    %\multicolumn{3}{l}{\textit{Multilingual transfer}} \\
    Multilingual & 40.16 & 54.35 \\
    Multilingual + TAPT & 37.68 & 59.25 \\
    \midrule
    \multicolumn{3}{l}{\textit{Transfer with selected sources}} \\
    TOP3 source fwd & 36.81 & 55.84 \\
    TOP3 source fwd + TAPT & 42.43 & 62.94 \\
    TOP3 source bwd & 41.79 & 66.77 \\
    TOP3 source bwd + TAPT & \textbf{43.50} & \textbf{67.32} \\
    \midrule
    \multicolumn{3}{l}{\textit{BERTweet with English translation}} \\
    BERTweet & 36.84 & 38.27 \\
    BERTweet + TAPT & 40.93 & 64.09 \\
    \midrule
    \multicolumn{3}{l}{\textit{Submitted systems}} \\
    mitchelldehaven & \textbf{46.23} & 66.96 \\
    UCAS  & 45.82 & 70.47 \\
    ymf924 & 45.34 & 70.39 \\
    UM6P  & 45.27 & 69.53 \\
    \textbf{NLNDE (ours)} & 44.97 & \textbf{70.86} \\
    \bottomrule
    \end{tabular}%
  \caption{Performance on subtask C: Zero-shot cross-lingual sentiment analysis. We experiment with (1) multilingual transfer, (2) transfer with selected sources, and (3) BERTweet with English translations. The model with TAPT and backward source language selection achieves the best overall results, which again demonstrates the effectiveness of the two approaches. The submission results come from the ensemble of different seeds of models with source selection and TAPT (forward and backward).}
  \label{tab:task_c_results}%
\end{table}%

The zero-shot cross-lingual transfer task is particularly challenging, especially when both the source and target languages are low-resource. In this subtask, we also employ different strategies: (1) multilingual transfer, (2) transfer with selected source and (3) BERTweet with English-translated samples.

First, we perform multilingual training, i.e., use all available training datasets from subtask A to fine-tune the AfroXLM-R model. We also perform task-adaptive pretraining with unlabeled multilingual texts here, as in the previous two subtasks. 

% We do not apply language-adaptive pretraining here due to time constraints and the fact that LAPT did not have a significant impact on the performance on top of TAPT (see Section \ref{sec:results_subtask_a}).

Second, we perform forward and backward source language selection for the cross-lingual transfer (as detailed in Section \ref{sec:source selection}). Here, we use the top 3 selected languages as the transfer source, as they show better performance than using all selected languages as the source in practice. We apply TAPT by using the unlabelled task-specific data from selected sources %languages and the target language.
and the target language.

Finally, we experiment with translating all tweets from the 14 languages to English using the \textit{pygoogletranslate} API.\footnote{https://github.com/Saravananslb/py-googletranslation} We investigate how the English BERTweet model performs for sentiment analysis. We also perform TAPT on the BERTweet model to adapt it to the unlabelled translated English dataset and then fine-tune the model with the labelled translated English dataset.

The results of subtask C are given in Table \ref{tab:task_c_results}. As in the previous subtasks, task-adaptive pretraining largely improves performance in all settings. Second, in 7 out of 8 cases, transfer learning with only selected source languages outperforms the multilingual counterparts trained on all languages. The model with a combination of TAPT and backward source language selection achieves the best overall results, which demonstrates the effectiveness of both strategies in subtask C.

Sentiment analysis based on English translations shows competitive performance, but still underperforms transfer learning with source selection. One possible reason could be that the translation quality is not good enough to accurately translate all relevant words with emotional meanings.
%, even though the general meaning can be retained. This also reflects the underdevelopment of African language translation research.

\subsection{Discussion}
\label{sec:results_discussion}
In summary, our work shows that adaptive pretraining and transfer learning with source language selection are effective approaches to tackle sentiment analysis in low-resource languages. Specifically, we demonstrate that (1) adaptive pretraining, especially task-adaptive pretraining, is generally effective across different subtasks and task settings, and (2) transfer learning with source language selection leads to better results than monolingual training. Using only source languages with positive transfer gains for training increases the available training data size on the one hand, and avoids interference from dissimilar languages on the other hand. Notably, forward and backward source selection outperform groupings based on language families in our multilingual experiments.

\section{Limitation and Future Work}
\label{sec:limitation}
One limitation of our work is that the forward and backward selection strategies require a lot of comparative experiments to determine if a candidate language has a positive or negative effect on the target language. For $N$ languages, we need to perform $N * N$ transfer experiments for the comparison (as described in Section \ref{sec:source selection}). How to automatically select source languages with little manual work is an interesting research question for future work.

Additionally, we found that forward and backward source selection produce different source language results and thus show different transfer scores. In our experiments, neither method completely outperformed the other. We have no conclusive answer to which method is better. Also, We have not conducted an in-depth study on the relationship between the selected sources for the target language and their linguistic correlation. This is another limitation of the current work that could be addressed in future research -- in particular when involving language experts.

\section{Related Work}

\paragraph{African language-centric PLMs.}
Large multilingual PLMs, such as mBERT \cite{bert} and XLM-R \cite{xlm-r} cover more than 100 languages for natural language processing tasks and exhibit good generalization abilities over a large number of languages. However, most of them include few African languages due to the lack of large open-source monolingual corpora~\cite{hedderich-etal-2021-survey}. Prior work developed African language-centric PLMs to address this under-representation. Among them, AfriBERTa \cite{afriberta} uses the RoBERTa \cite{roberta} architecture and trains the model from scratch with corpora from 11 African languages. AfroLM \cite{afro-lm} proposes to use a novel self-active learning framework and the model is trained from scratch on 23 African languages. Another strategy is to initialize a language model from an existing model and continue to train it with a special focus on African languages. AfroXLM-R \cite{afroxlm-r} performs multilingual adaptive fine-tuning based on XLM-R on 17 highest-resourced African languages and 3 other high-resource languages spoken on the African continent. AfroXLM-R performs well on African language tasks, such as named entity recognition 
%(NER) 
and sentiment analysis \cite{afroxlm-r, afro-lm}. We therefore use it as the base model in this shared task.

\paragraph{Adaptive pertaining.}
\newcite{dont-stop} demonstrate that it is helpful to further tailor a pretrained model to a target domain and task. In particular, they introduce domain-adaptive pretraining, which continues the pretraining of the model on domain-specific unlabeled data, and task-adaptive pretraining, which further pretrains the model on the task’s unlabeled data. Experimental results show that these two strategies lead to remarkable performance gains. We adopt this idea to our tasks. %By applying language-adaptive and task-adaptive pretraining, the performance is effectively improved across all subtasks and task settings.

\paragraph{Transfer learning with source selection.}
Selecting data for transfer learning has been explored in different prior work, i.e., \newcite{ruder-plank, lin-etal-2019-choosing, 10.1093/bioinformatics/btac297}. For example, \newcite{ruder-plank} learn to select positive sources using Bayesian optimization. %and evaluate the approach across models, domains and tasks. 
LangRank \cite{lin-etal-2019-choosing} considers the source language selection for transfer learning as a ranking problem. They train a ranking model to select languages with a positive transfer gain from a larger set of possible languages. In contrast, we adopt the idea of forward and backward feature selection \cite{fwd-bwd-1} and use a much simpler approach based on transfer score comparison to select source languages.

\section{Conclusion}

In this work, we introduce our sentiment analysis system for the AfriSenti shared task, which is ranked first in 8 out of 15 tracks and performs competitively on the others. It consists of language-adaptive and task-adaptive pretraining on top of the AfroXLM-R model, together with transfer learning with source language selection. We demonstrate that tailoring the pretrained model to the target language and task considerably improves the performance across all task settings. Additionally, transfer learning with source language selection further improves the results in the multilingual and zero-shot cross-lingual tasks by avoiding potential negative transfer gains from dissimilar languages. A future research direction is to automatically select source languages with positive transfer gains without the need of manually comparing the source-to-target transfer score.

\section*{Acknowledgments}
We thank the AfriSenti organizers for their time to prepare the data for a large variety of languages and run the competition in a smooth way. The shared task provides an excellent platform for researchers to collaborate and share their knowledge, and also promotes the future development in the field of NLP for African languages. 

%

\bibliography{anthology,custom}

\clearpage
\appendix

\section{Appendix}

\subsection{Results on \textit{dev set*}}
\label{sec:result_dev_set}
Here, we provide the experimental results of all three subtasks on our processed \textit{dev set*} (for details, please refer to Section \ref{sec:data_preprocess}).

\begin{table*}[htbp]
  \centering
  \footnotesize
  \scalebox{0.9}{
    \begin{tabular}{lccccccccccccc}
    \toprule
    \multicolumn{1}{c}{\textbf{Model}} & \textbf{am} & \textbf{dz} & \textbf{ha} & \textbf{ig} & \textbf{kr} & \textbf{ma} & \textbf{pcm} & \textbf{pt} & \textbf{sw} & \textbf{ts} & \textbf{twi} & \textbf{yo} & \textbf{overall} \\
    \midrule
    AfroXLM-R & 45.20 & 58.63 & 80.27 & 80.97 & 70.77 & 59.43 & 51.83 & 47.10 & 63.27 & 54.40 & 35.97 & 77.00 & 66.19 \\
    \midrule
    \multicolumn{14}{l}{\textit{AfroXLM-R with adaptive pretraining}} \\
    LAPT  & 56.84 & 66.96 & 79.72 & 82.06 & 71.76 & 74.74 & 72.50 & 68.24 & 62.42 & \textbf{64.06} & 52.84 & 77.34 & 73.67 \\
    TAPT  & 62.06 & \textbf{71.38} & \textbf{80.96} & 82.30 & 71.32 & \textbf{80.58} & \textbf{74.32} & \textbf{71.12} & \textbf{63.44} & 51.10 & \textbf{63.58} & 78.00 & 75.01 \\
    LAPT + TAPT & \textbf{63.42} & 71.12 & 80.36 & \textbf{82.72} & \textbf{73.20} & 79.44 & 74.06 & 69.04 & 62.14 & 63.78 & 63.56 & \textbf{78.14} & \textbf{75.13} \\
    \bottomrule
    \end{tabular}%
    }
  \caption{Subtask A (Monolingual sentiment analysis) results on \textit{dev set*}. Same as in Section \ref{sec:results_subtask_b}, TAPT achieves the best overall F1 score.}
  \label{tab:task_a_results_dev}%
\end{table*}%

\begin{table}[htbp]
\footnotesize
  \centering
    \begin{tabular}{lc}
    \toprule
    \multicolumn{1}{c}{\textbf{Model}} & \textbf{Overall} \\
    \midrule
    \multicolumn{2}{l}{\textit{Multilingual training}} \\
    Multilingual & 51.64 \\
    + TAPT & 73.64 \\
    \midrule
    \multicolumn{2}{l}{\textit{Monolingual training}} \\
    Monolingual & 66.19 \\
    + LAPT & 73.67 \\
    + TAPT & 75.01 \\
    + LAPT \& TAPT & 75.13 \\
    \midrule
    \multicolumn{2}{l}{\textit{transfer with selected sources}} \\
    Fwd source transfer  & 71.39 \\
    Fwd source transfer + TAPT & \textbf{75.53} \\
    Bwd source transfer  &  71.37 \\
    Bwd source transfer + TAPT & 75.49 \\
    \bottomrule
    \end{tabular}%
  \caption{Subtask B (Multilingual sentiment analysis) results on \textit{dev set*}. The model with TAPT and forward source language selection achieves the best overall results with an F1 score of 75.53.}
  \label{tab:task_b_results_dev}%
\end{table}%

\begin{table}[htbp]
\footnotesize
  \centering
    \begin{tabular}{lcc}
    \toprule
    \multicolumn{1}{c}{\textbf{Model}} & \textbf{or} & \textbf{tg} \\
    \midrule
    \multicolumn{3}{l}{\textit{Multilingual transfer}} \\
    Multilingual & 48.16 & 50.94 \\
    Multilingual + TAPT & 52.20 & 57.14 \\
    \midrule
    \multicolumn{3}{l}{\textit{Transfer with selected sources}} \\
    TOP3 source fwd & 49.00 & 56.96 \\
    TOP3 source fwd + TAPT & \textbf{57.90} & 61.83 \\
    TOP3 source bwd & 52.42 & 62.88 \\
    TOP3 source bwd + TAPT & 57.22 & \textbf{63.48} \\
    \midrule
    \multicolumn{3}{l}{\textit{BERTweet with English translation}} \\
    BERTweet & 50.12 & 54.10 \\
    BERTweet + TAPT & 56.47 & 63.43 \\
    \bottomrule
    \end{tabular}%
  \caption{Subtask C (Zero-shot cross-lingual sentiment analysis) results on \textit{dev set*}. Models with TAPT and source selection achieves the best overall results with 57.90 F1 score on \textit{or} and 63.48 F1 score on \textit{tg}.}
  \label{tab:task_c_results_dev}%
\end{table}%

\end{document}